%% file: main.tex
\documentclass[letterpaper, 10 pt, conference]{ieeeconf}
\IEEEoverridecommandlockouts
\input{pre}

\usepackage{cite}
\usepackage{amsmath,amssymb,amsfonts}
\usepackage{algorithmic}
\usepackage{graphicx}
\usepackage{textcomp}
\usepackage{xcolor}

\begin{document}

\title{
\textbf{TWIST{\color{deepred2}2}}\\
Scalable, Portable, and Holistic Humanoid Data  Collection System
}

\author{
Yanjie Ze$^{12}$\quad Siheng Zhao$^{13}$\quad Weizhuo Wang$^{12}$\\ Angjoo Kanazawa$^{14\dag{}}$\quad Rocky Duan$^{1\dag{}}$\quad Pieter Abbeel$^{14\dag{}}$\quad Guanya Shi$^{15\dag{}}$\quad Jiajun Wu$^{2\dag{}}$\quad C. Karen Liu$^{12\dag{}}$\\
$^1$Amazon FAR\quad $^2$Stanford University\quad  $^3$USC \quad  $^4$UC Berkeley \quad  $^5$CMU\quad $^{\dag{}}$Equal Advising
}

\twocolumn[{%
\renewcommand\twocolumn[1][]{#1}%
\maketitle
\vspace{-0.3in}
\begin{center}
    \centering
    \captionsetup{type=figure}
    \includegraphics[width=1.0\linewidth]{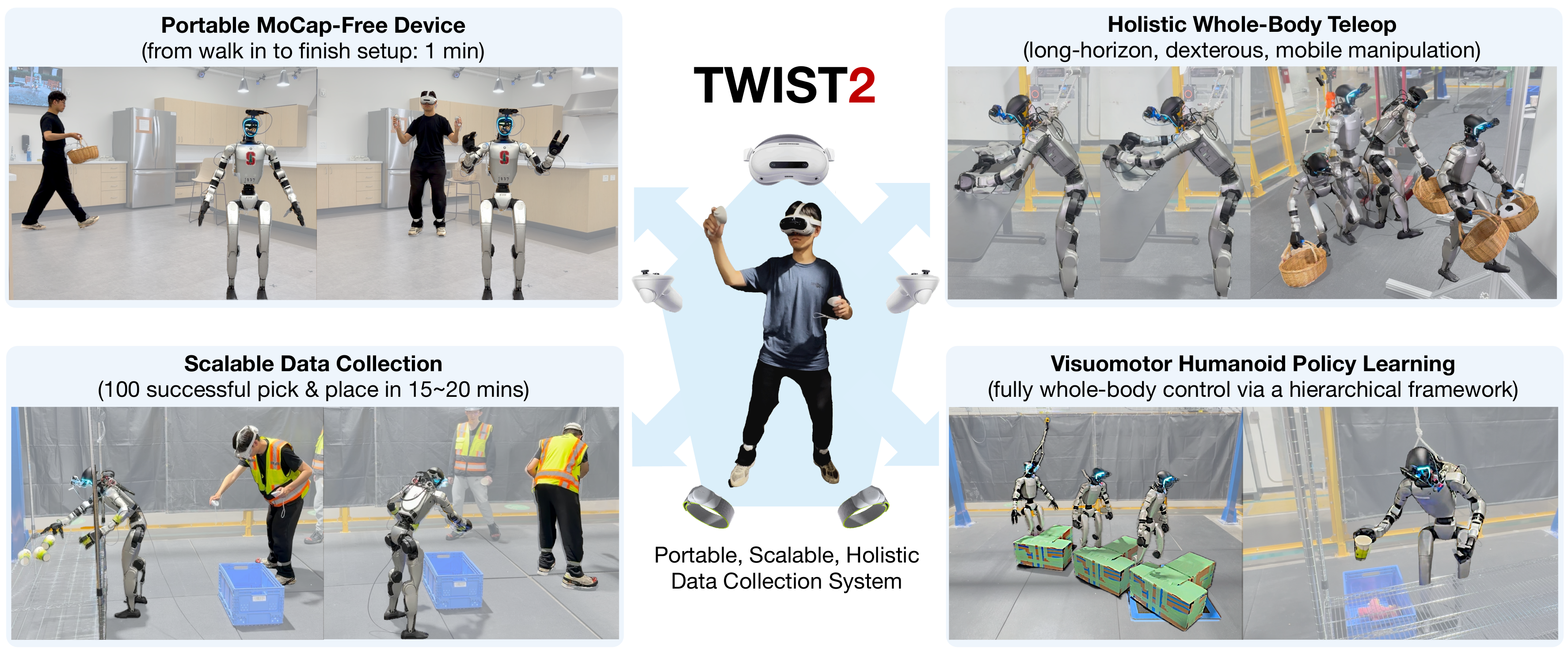}
    \caption{We introduce \ours, a holistic humanoid data collection system designed with scalability and portability. \ours enables scalable data collection, fast setup, and enjoyable user experience compared to MoCap solutions such as TWIST~\cite{ze2025twist}, while maintaining the full whole-body control. We build a 2-DoF Neck (\ours Neck) to enable egocentric teleoperation, which costs  \$250.
    With \ours, our robots are able to perform long-horizon, dexterous, mobile whole-body manipulation and legged manipulation. All tasks are achieved with streamed robot egocentric vision, full whole-body control, and a single operator. We further train visuomotor policies upon data collected via \ours. Our entire system is open-sourced at \ourwebsite and ensure full reproducibility. }
    \label{fig:teaser}
\end{center}
\vspace{-0.05in}
}]

\begin{abstract}
\footnotetext[1]{Work done during the internship of Yanjie Ze, Siheng Zhao, and Weizhuo Wang at Amazon Frontier AI \& Robotics (FAR).}
Large-scale data has driven breakthroughs in robotics, from language models to vision-language-action models in bimanual manipulation. However, humanoid robotics lacks equally effective data collection frameworks. Existing humanoid teleoperation systems either use decoupled control or depend on expensive motion capture setups. We introduce \ours, a portable, mocap-free humanoid teleoperation and data collection system that preserves full whole-body control while advancing scalability. Our system leverages PICO4U VR for obtaining real-time whole-body human motions, with a custom 2-DoF robot neck (cost around \$250) for egocentric vision, enabling holistic human-to-humanoid control. We demonstrate long-horizon dexterous and mobile humanoid skills and  we can collect 100 demonstrations in 15 minutes with an almost 100\% success rate. Building on this pipeline, we propose a hierarchical visuomotor policy framework that autonomously controls the full humanoid body based on egocentric vision. Our visuomotor policy successfully demonstrates whole-body dexterous manipulation and dynamic kicking tasks. The entire system is fully reproducible and open-sourced at \ourwebsite. Our collected dataset is also open-sourced at \href{https://twist-data.github.io}{\color{deepred}https://twist-data.github.io}.

\end{abstract}

\input{sec/1_intro}

\input{sec/2_related}

\input{sec/3_method}

\input{sec/4_exp}

\input{sec/5_conclu}
\bibliographystyle{IEEEtran}
\bibliography{main}

\end{document}

%% file: pre.tex
\usepackage{multicol}
\usepackage{graphicx}
\usepackage{xspace}
\usepackage{xcolor}
\usepackage{caption}
\usepackage{amsmath,amssymb} 
\usepackage{pifont}

\usepackage{wrapfig}
\usepackage{bbding}  
\usepackage{pifont}  
\usepackage[bookmarks=true]{hyperref}

\usepackage{subcaption} 

\usepackage{booktabs}
\usepackage{float}
\usepackage{amsmath}  
\usepackage{amssymb}  
\usepackage{multirow, booktabs}
\usepackage{siunitx}
\usepackage{dcolumn} 
\newcolumntype{d}{D{/}{/}{-1}} 
\newcommand{\ours}[0]{TWIST\textbf{2}\xspace}

\newcommand{\ourwebsite}[0]{\href{https://yanjieze.com/TWIST2}
{\color{deepred}https://yanjieze.com/TWIST2}\xspace}

\definecolor{deepgreen}{RGB}{63, 126, 49}
\definecolor{deepred2}{RGB}{196, 49, 25}

\newcommand{\cmark}{\textcolor{deepgreen}{\ding{51}}}%
\newcommand{\xmark}{\textcolor{deepred2}{\ding{55}}}%

\usepackage{colortbl}
\definecolor{ourcolor}{HTML}{99e0eb}
\definecolor{ourblue}{HTML}{27a2c3}

\definecolor{tablecolor}{HTML}{ccf2f5} 

\definecolor{tablecolor2}{HTML}{ffcdb4}
\definecolor{citecolor}{HTML}{fe7b5b}
\definecolor{grey}{rgb}{0.9, 0.9, 0.9}
\usepackage{amssymb}

\usepackage{listings}
\definecolor{gred}{rgb}{0.859,0.267,0.216}
\definecolor{ggreen}{rgb}{0.059,0.616,0.345}

\definecolor{deepblue}{HTML}{27a2c3}

\definecolor{deepred}{HTML}{7c2320}

%% file: sec/1_intro.tex
\section{Introduction}
\input{tables/comparison_control_v2}

The transformative power of large-scale data has fundamentally reshaped machine learning, driving breakthrough achievements from large language models like GPT-4~\cite{openai2023gpt4} to the recent success of vision-language-action (VLA) models in robotics.
In the realm of bimanual manipulation, models such as $\pi_0$~\cite{black2024pi_0} and $\pi_{0.5}$~\cite{physicalintelligence2025pi05} have demonstrated unprecedented capabilities, directly enabled by the robust and scalable data collection infrastructure~\cite{zhao2023aloha,aloha22024,wu2023gello}.
However, this data-driven revolution has yet to reach humanoid robots, where the absence of equally effective data collection frameworks continues to limit progress toward human-level versatile manipulation and locomotion.

As summarized in Table~\ref{tab:related work comparison}, existing humanoid teleoperation systems fall into three broad categories: a) \textit{Decoupled control} of lower and upper body (e.g., MobileTV~\cite{lu2024mobiletv}, HOMIE~\cite{ben2025homie}); b) \textit{Partial whole-body control} that coordinates selected body segments such as arms and torso while legs track base velocity commands (e.g., AMO~\cite{li2025amo}, CLONE~\cite{li2025clone}); c) \textit{Full whole-body control} that directly tracks human body pose across all joints including arms, torso, and legs in a unified manner (e.g., HumanPlus~\cite{fu2024humanplus}, TWIST~\cite{ze2025twist}).
Among these, VR-based solutions such as AMO and CLONE offer practicality but are limited to mobile skills with simple locomotion, falling short of capturing dynamic whole-body coordination skills that humans naturally exhibit.
In contrast, full whole-body control holds the greatest promise for unleashing the versatility of humanoid robots, as evidenced by TWIST~\cite{ze2025twist}.
However, such systems typically depend on expensive, non-portable motion capture setups, restricting deployment to lab environments.

In this work, we introduce \ours, a humanoid teleoperation and data collection system that preserves the power of full whole-body control while advancing portability and scalability.
Our design leverages PICO4U~\cite{pico4ultra_2023}, a lightweight VR device that provides whole-body motion streaming using a head goggle, handheld controllers, and two motion trackers on the ankles, without requiring expensive motion capture systems.
Recognizing that egocentric vision is crucial for human-like task execution, we design a low-cost and non-invasive neck that seamlessly integrates with Unitree G1 and our VR teleoperation ecosystem.
With these portable components, we build a comprehensive retargeting pipeline from full human body poses of PICO to corresponding humanoid motor joint positions. 
To execute the retargeted motions on the robot, we train a robust motion tracking controller using reinforcement learning and large-scale simulation interaction on carefully curated motion data.

These elements together enable efficient, long-horizon, in-the-wild teleoperation and data collection without reliance on motion capture systems, and only requiring a single operator.
We showcase that 1) we can teleoperate robots to perform very long-horizon and fine-grained whole-body dexterous skills such as folding towels and mobile skills such as transporting objects through the door, and 2) we can collect human demonstrations efficiently, \textit{e.g.}, collecting around 100 successful demonstrations in 20 minutes without failure. We also find that egocentric active stereo vision is essential for the long-horizon mobile and dexterous teleoperation.

Building on this scalable data collection pipeline, we further propose a hierarchical visuomotor policy learning framework consisting of two components.
The first component is the same motion tracking controller used during teleoperation, which serves as a low-level controller.
The second component is a Diffusion Policy that directly predicts whole-body joint positions based on visual observations that feeds into the low-level controller.
To our knowledge, this is the first policy learning framework that enables vision-based autonomous control of the full humanoid body, moving beyond simplified commands such as root velocity. Importantly, this capability is made possible by our data collection system, which provides the high-quality demonstrations needed for training. 

We showcase a few representative results where our humanoid robot autonomously performs a) consecutive whole-body dexterous pick \& place and b) continuous kicking of a T-shaped box to target regions (Kick-T), illustrating the potential of this new framework.



To summarize, our main contributions are:
\begin{enumerate}
    \item A portable, mocap-free humanoid teleoperation and data collection system with full whole-body control, enhanced with an attachable neck for egocentric active vision.
    
    \item A hierarchical whole-body visuomotor policy learning framework that achieves full whole-body control.
    \item Demonstration of long-horizon teleoperation skills such as towel folding/unfolding and object transporting through the door, effective data collection, and new autonomous humanoid skills including whole-body dexterous pick \& place and Kick-T.
\end{enumerate}
\textbf{Our system, data, and model are fully open-sourced at \ourwebsite to ensure full reproducibility.
}

%% file: tables/comparison_control_v2.tex
\begin{table*}[ht]
\centering
\caption{\textbf{Comparison of recent humanoid data collection systems}. We compare existing humanoid teleoperation systems across key dimensions essential for effective data collection. \ours is the first system to combine full whole-body control with portability, achieving comprehensive capabilities including egocentric teleoperation, accurate tracking, and single-operator efficiency. Unlike previous works that either sacrifice portability for full whole-body control (TWIST) or sacrifice full whole-body control for portability (AMO, CLONE), our system achieves all critical requirements for scalable humanoid data collection.}
\label{tab:related work comparison}
\resizebox{1.0\textwidth}{!}{%
\begin{tabular}{l|l|cccc|ccccc}
\toprule
\textbf{Humanoid Data}  & 
& \multicolumn{4}{c}{\textbf{Portability \& Scalability}} 
& \multicolumn{4}{|c}{\textbf{Holistic Control}}  \\

\textbf{Collection System} & \textbf{Category} & \textbf{Source} 
& \textbf{Portable} & \textbf{No Calibration} & \textbf{Single Operator} 
& \textbf{Whole-Body Tracking} & \textbf{Egocentric Teleop} 
& \textbf{Foot Control} & \textbf{Wrist Control} \\

\midrule

HOMIE~\cite{ben2025homie} & Decoupled & Exoskeleton & \xmark & \cmark & \cmark & \xmark & \xmark & \xmark & \cmark \\
AMO~\cite{li2025amo} & Partial & VR & \cmark & \cmark & \xmark & \xmark & \cmark & \xmark & \cmark \\
CLONE~\cite{li2025clone} & Partial & VR & \cmark & \cmark & \cmark & \xmark & \xmark & \xmark & \cmark \\
TWIST~\cite{ze2025twist} & Full & MoCap & \xmark & \xmark & \cmark & \cmark & \xmark & \cmark & \xmark \\

\textbf{\ours} (ours) & Full & VR & \cmark & \cmark & \cmark & \cmark & \cmark & \cmark & \cmark \\

\bottomrule
\end{tabular}%
}
\end{table*}

%% file: sec/2_related.tex
\section{Related Work}

\subsection{Whole-Body Humanoid Teleoperation}

Teleoperation is crucial for enabling humanoid robots to interact with complex real-world environments and perform sophisticated loco-manipulation tasks.
Unlike wheel-based robots or tabletop arms, the anthropomorphic nature of humanoids makes whole-body control the most natural and effective teleoperation approach~\cite{fu2024humanplus,he2024omnih2o,ze2025twist,li2025clone,cheng2024tv,ze2024idp3,li2025amo}.
As shown in Table~\ref{tab:related work comparison}, we categorize recent works into three categories: a) decoupled control, b) partial whole-body control, and c) full whole-body control.
Full whole-body control, as demonstrated by TWIST~\cite{ze2025twist}, shows promising results in coordinated whole-body dexterity, which is the primary focus of this work.
As detailed in Table~\ref{tab:related work comparison}, we identify several critical aspects in scalable \& holistic teleoperation and data collection that remain lacking in previous works for real-world deployment. which we address comprehensively in this work.


\begin{figure*}
    \centering
\includegraphics[width=1.0\linewidth]{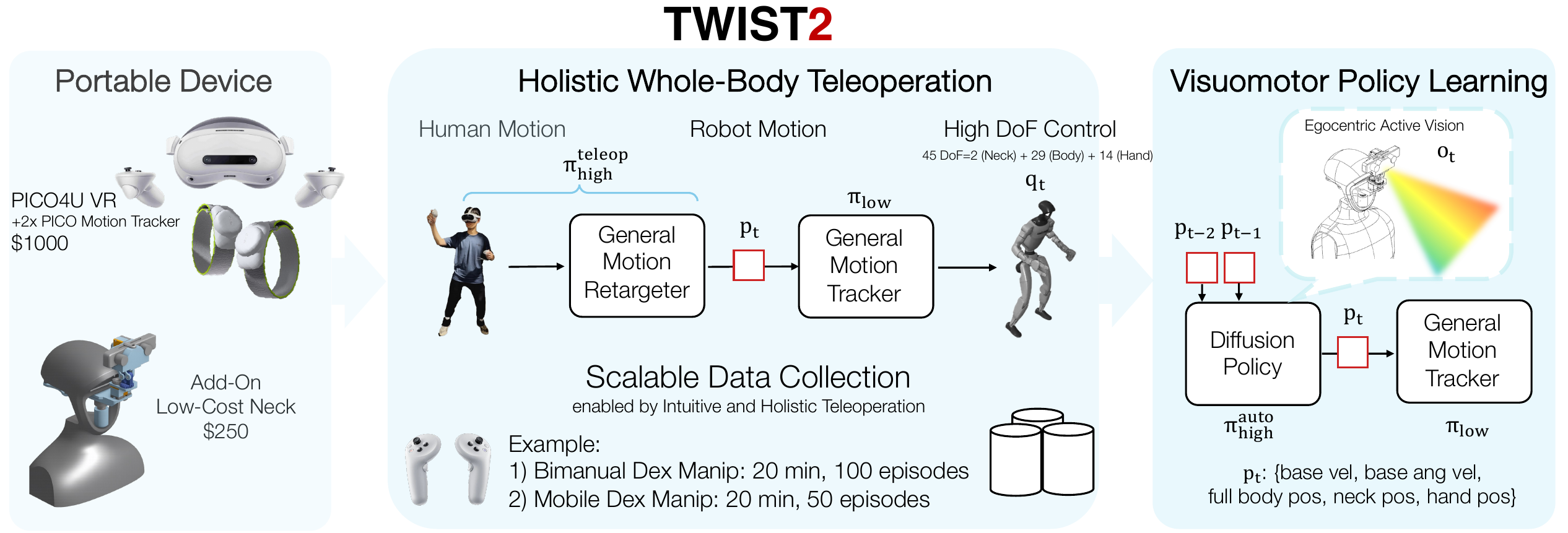}
    \caption{System overview of \ours. We build a holistic humanoid teleoperation system with portable devices and egocentric active vision, enabling scalable imitation data collection. With data collected, we build a hierarchical visuomotor policy learning framework that directly predicts whole-body joint positions.
    }
    \label{fig:system overview}
    \vspace{-0.2in}
\end{figure*}

\subsection{Visual Humanoid Control}

Previous works on visual humanoid control predominantly rely on LiDAR for perceptive locomotion~\cite{wang2025beamdojo,videomimic,long2024learninghumanoidlocomotionperceptive}, typically employing task-specific sim-to-real reinforcement learning (RL) approaches. 
Recent works like HEAD~\cite{chen2025head} propose keypoint-based hierarchical frameworks with humanoid egocentric vision, while limiting in simple navigation tasks. VideoMimic~\cite{videomimic} introduced a real2sim2real pipeline that enables real robots to perform environment interactions such as sitting, though their interactions remain limited to static settings like the ground or stone chairs. 
Some works such as PDC~\cite{luo2025pdc} are conducted only in simulation and face significant sim-to-real transfer challenges.
In contrast, our work focuses on developing general visuomotor humanoid policies that can interact with complex environments and perform long-horizon whole-body loco-manipulation and legged manipulation tasks—capabilities not demonstrated in previous works.


%% file: sec/3_method.tex
\section{Our System}
We introduce \ours, a scalable, portable, and holistic humanoid teleoperation and data collection system (see Figure~\ref{fig:teaser} for capabilities). As illustrated in Figure~\ref{fig:system overview}, our system consists of four main components: a humanoid robot equipped with active vision (Section~\ref{sec:Humanoid Robot with Active Vision}), portable motion capture using VR devices (Section~\ref{sec:portable human data source}), holistic human-to-robot motion retargeting (Section~\ref{sec:holistic human-to-humanoid retargeting}), a general motion tracker for low-level control (Section~\ref{sec:training controller}). These components work together to enable scalable data collection (Section~\ref{sec:scalable humanoid data collection}) and autonomous visuomotor policy execution (Section~\ref{sec: whole-body policy learning}).

\subsection{Problem Formulation}

We focus on enabling humanoid robots to perform diverse whole-body dexterous tasks with their own egocentric vision and proprioception within a single unified framework.
To this end, we propose a two-level hierarchical control framework, consisting of a low-level controller $\pi_{\text{low}}$ and a high-level controller $\pi_{\text{high}}$.

\noindent\textbf{Low-level control.} 
We formulate the low-level controller $\pi_{\text{low}}$ as a \textit{general motion tracking} problem, so that our low-level control is task-agnostic. At each timestep, the low-level controller receives a reference command vector composed of root translational velocity in the $x$ and $y$ axes, root $z$ position, root roll/pitch angles, root yaw angular velocity, and whole-body joint positions:
\begin{equation}
\mathbf{p}_{\text{cmd}}=\Big[\, \dot{x}_{\text{ref}},\; \dot{y}_{\text{ref}},\; z_{\text{ref}},\; 
\phi_{\text{ref}},\; \theta_{\text{ref}},\; \dot{\psi}_{\text{ref}},\; \mathbf{q}_{\text{ref}} \,\Big].
\end{equation}
In addition, it has access to robot proprioception, including root orientation and angular velocity from IMU readings, as well as joint positions and velocities from encoders:
\begin{equation}
\mathbf{s} = \Big[\, \omega,\; \dot{\omega},\; \mathbf{q},\; \dot{\mathbf{q}} \,\Big].
\end{equation}
The controller outputs desired joint positions,
\begin{equation}
\mathbf{q}_{\text{tgt}} = \pi_{\text{low}}(\mathbf{s}, \mathbf{p}_{\text{cmd}}),
\end{equation}
at 50Hz, which are then tracked by a PD controller to generate the final torque:
\begin{equation}
\tau = K_{\text{P}} \, (\mathbf{q}_{\text{tgt}} - \mathbf{q}) - K_{\text{D}} \, \dot{\mathbf{q}}.
\end{equation}

\noindent\textbf{High-level control.} 
The high-level controller $\pi_{\text{high}}$ focuses on generating task-specific motion commands $\mathbf{p}_{\text{cmd}}$ conditioned on egocentric vision. We have two variants in this work: (1) a teleoperation policy $\pi_{\text{high}}^{\text{teleop}}$, and (2) a visuomotor policy $\pi_{\text{high}}^{\text{auto}}$. Both map visual observations $\mathbf{o}$ and proprioceptive states $\mathbf{s}$ into commands:
\begin{equation}
\mathbf{p}_{\text{cmd}} = \pi_{\text{high}}(\mathbf{o}, \mathbf{s}).
\end{equation}
In this work, we employ $\pi_{\text{high}}^{\text{teleop}}$, \textit{i.e.}, the human teleoperator plus the motion retargeter, to collect observation–action pairs $(\mathbf{o}, \mathbf{s}, \mathbf{p}_{\text{cmd}})$, which are then used to train $\pi_{\text{high}}^{\text{auto}}$, \textit{e.g.}, a Diffusion Policy. 

\noindent\textbf{Interface design.} 
There are two key aspects of our command interface $\mathbf{p}_{\text{cmd}}$: 
(1) We use relative root translations/rotations rather than absolute poses, so that our system does not rely on accurate global state estimation~\cite{truong2025beyondmimic}, and remains stable during very long-horizon operation; 
(2) We include whole-body joint positions instead of simplifying lower-body control as root velocity only~\cite{li2025amo,li2025clone,lu2024mobiletv}, which enables more precise control of lower-body movements and unlocks tasks such as legged manipulation and dancing.

\subsection{Humanoid Robot with Active Vision}
\label{sec:Humanoid Robot with Active Vision}
We use Unitree G1 with 29 DoF (3 DoF waist + two 6 DoF legs + two 7 DoF arms), equipped with two 7 DoF Dex31 hands. We find that neck DoFs are essential for effective and long-horizon teleoperation, so we build a portable robot neck with yaw and pitch DoFs.

\begin{figure}[htbp]
    \centering
    \vspace{-0.1in}
\includegraphics[width=0.8\linewidth]{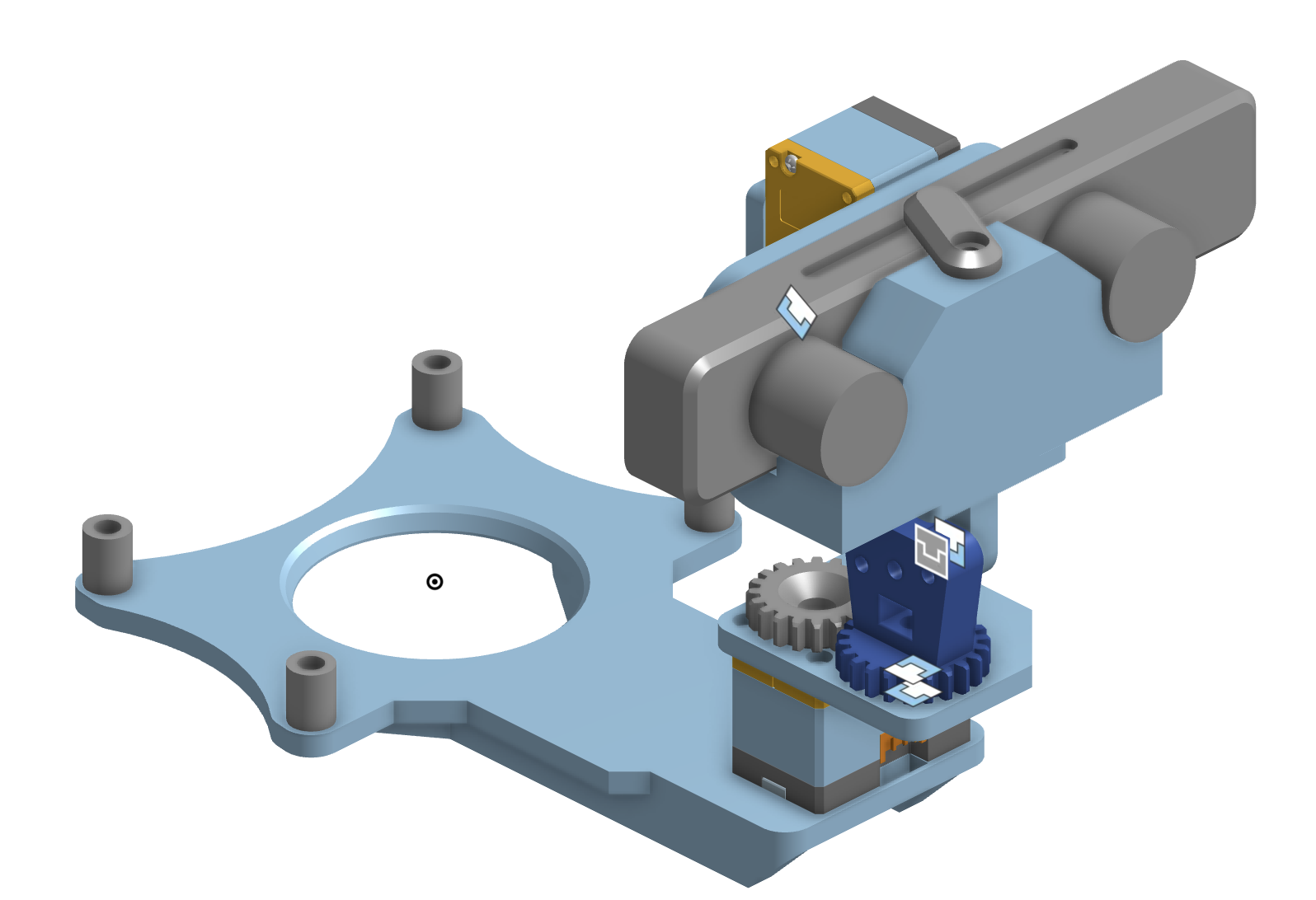}
    \caption{\ours Neck.  We design a simple yet effective 2-DoF neck that can be easily assembled for a non-expert user and can be attached/detached to/from a Unitree G1 without removing the original LiDAR.}
    \label{fig:neck design}
\end{figure}
\begin{figure}[htbp]
    \centering
    \vspace{-0.1in}
\includegraphics[width=0.8\linewidth]{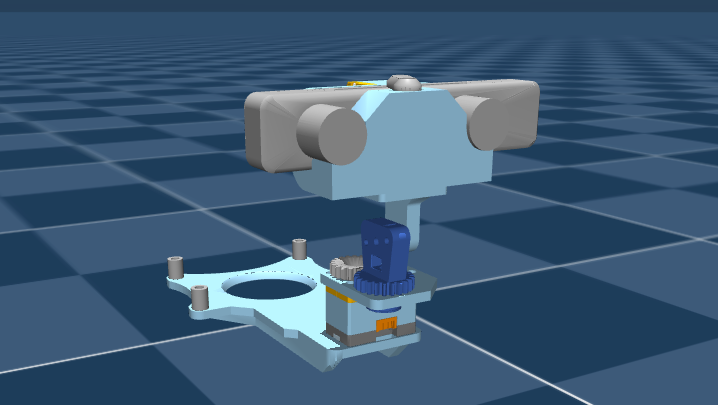}
    \caption{\ours Neck in MuJoCo. To facilitate the research in simulation and standardize our data, we  build MuJoCo XML files for our TWIST2 neck. }
    \label{fig:neck in mujoco}
\end{figure}
\begin{figure}[htbp]
    \centering    \includegraphics[width=0.8\linewidth]{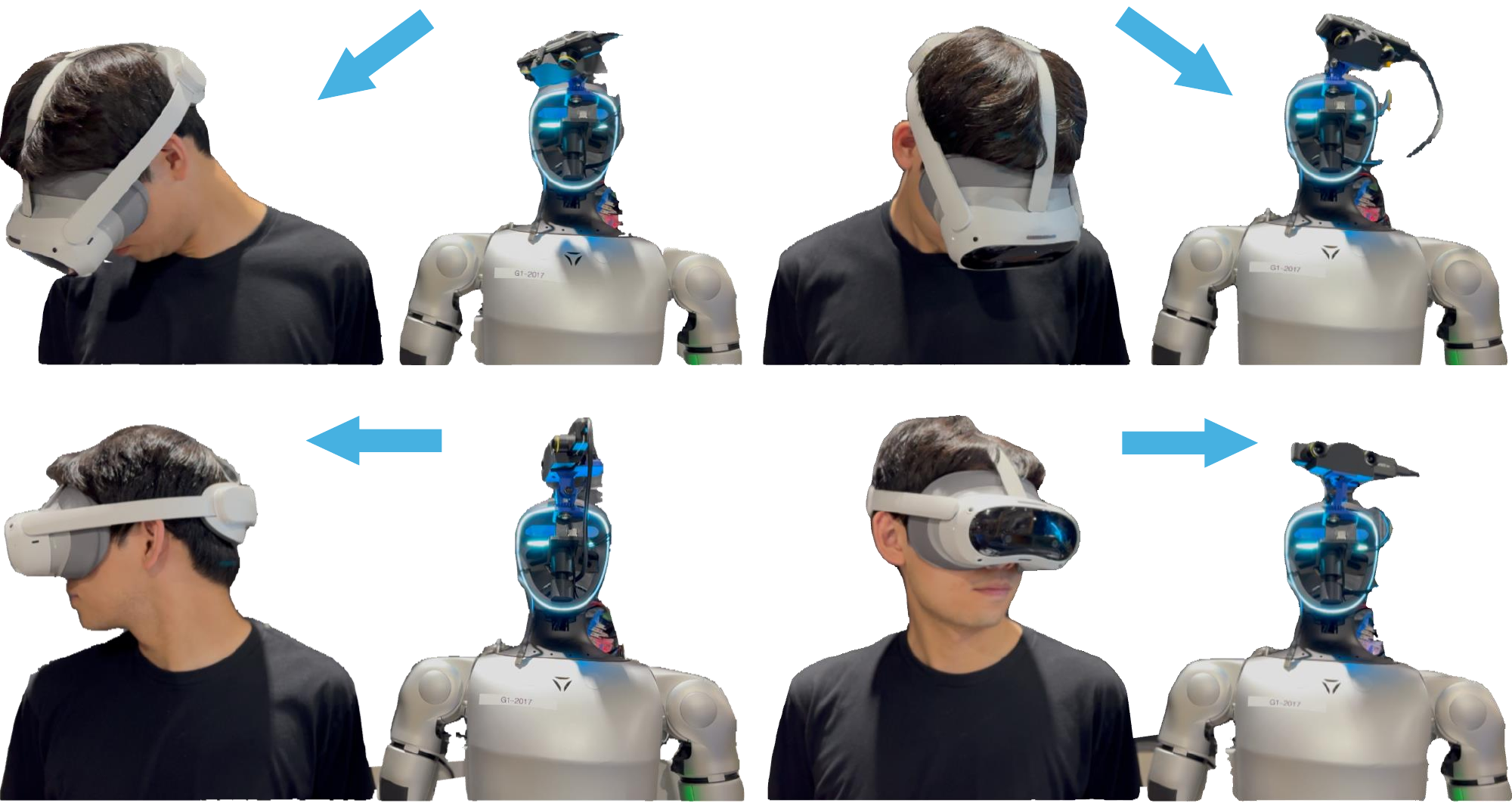}
    \caption{Mimic the human neck with the robot neck. We found that a 2 DoFs neck (yaw and pitch) is sufficient to mimic major human neck movements.}
    \label{fig:neck}
    \vspace{-0.2in}
\end{figure}

\noindent\textbf{Add-on low-cost neck (\ours Neck).} Unlike recent works~\cite{li2025amo,xiong2025via} that build built-in necks, we design an add-on neck module that can be seamlessly attached to the Unitree G1 without disassembling its original head (see Figure~\ref{fig:neck design}). Our design is inspired by ToddlerBot~\cite{shi2025toddlerbot}. We use two Dynamixel XC330-T288 motors to control the yaw and pitch angles, connected via a U2D2 and powered by the onboard 12V/5A supply. All structural parts are 3D printed. The cost of the neck is \$250. We use Zed Mini as our stereo camera attached to the neck (the ZED Mini stereo camera will cost extra \$400). Since human roll DoF is rarely used in everyday interaction, we find that the two-DoF design already enables smooth and human-like neck motions (Figure~\ref{fig:neck}). To further standardize the TWIST2 neck usage, we build the corresponding simulation model in MuJoCo as shown in Figure~\ref{fig:neck in mujoco}.

\subsection{Portable MoCap-Free Whole-Body Human Data Source}
\label{sec:portable human data source}

To obtain real-time full human body poses in a portable manner, we utilize PICO 4U~\cite{pico4ultra_2023} combined with two
PICO Motion Trackers~\cite{pico_motion_tracker_2023} that are bound on the humans' calves to obtain global translations and rotations for each human body parts. Though PICO supports more than 2 motion trackers, we find the 2-tracker mode provides a more stable pose estimation.  The cost for such a setup is around \$1000. much cheaper and practical compared to an optical MoCap system. We use XRoboToolkit~\cite{zhao2025xrobotoolkit} for access to motion streaming from PICO (Figure~\ref{fig:VR human body}). The motion can be streamed at 100Hz. Notably, PICO does not require heavy calibration compared to the MoCap system. As shown in Figure~\ref{fig:teaser}, it takes around only 1 minute to finish the setup of PICO.

Compared to HTC Vive Tracker~\cite{htc_vive_tracker_3.0} that is used in recent demos of Boston Dynamics~\cite{BD2025_LBM_Atlas}, PICO's whole-body estimation does not require extra third-person view camera setup, thus more flexible.


\subsection{Holistic Human-to-Humanoid  Retargeting}
\label{sec:holistic human-to-humanoid retargeting}
In this section, we describe how human motion data is holistically leveraged to control the humanoid robot’s body, hands, and neck.

\begin{figure}[htbp]
    \centering
    \includegraphics[width=1.0\linewidth]{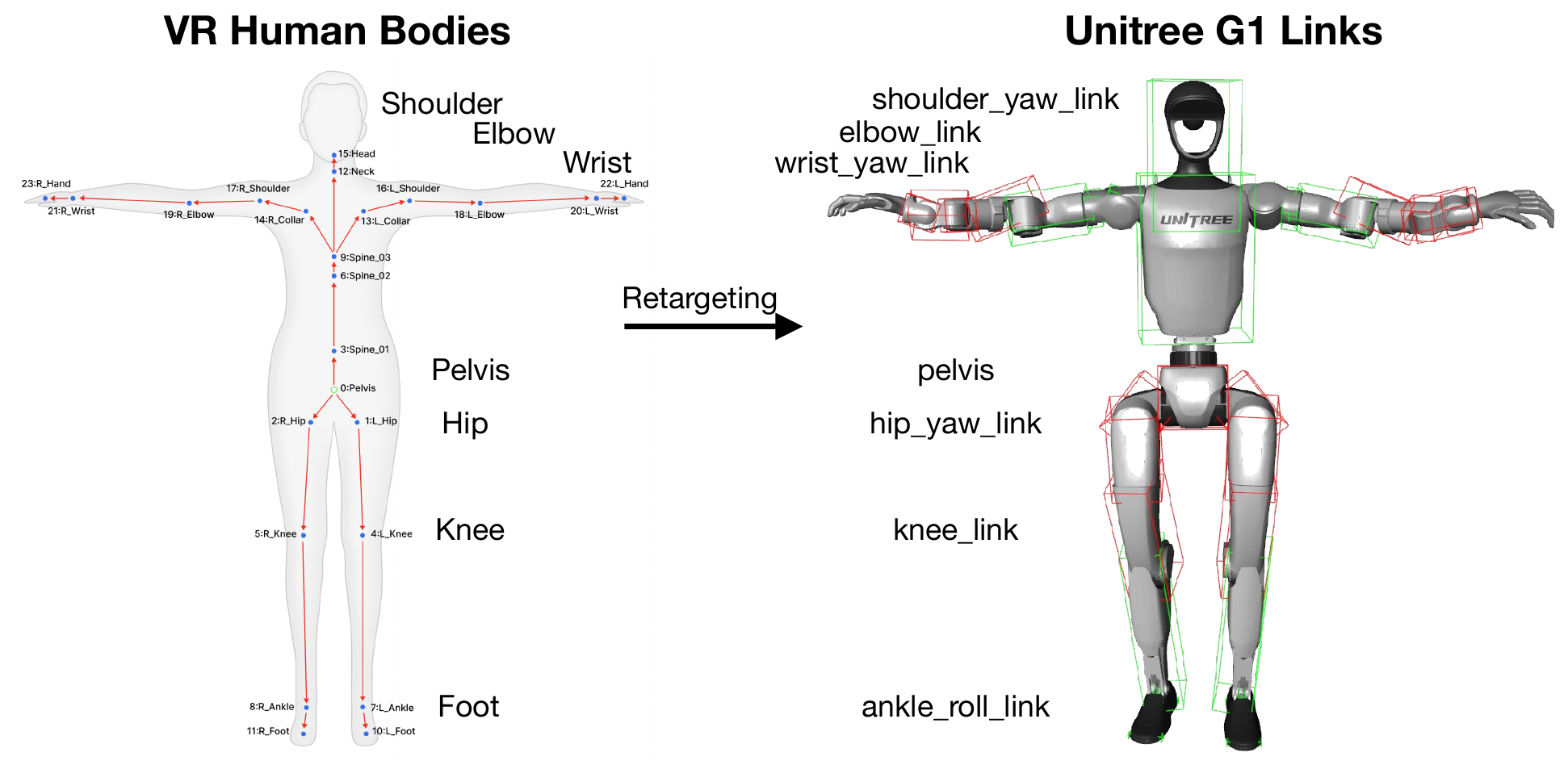}
    \caption{Mapping VR human bodies to robot links.}
    \label{fig:VR human body}
    \vspace{-0.1in}
\end{figure}

\noindent\textbf{Body retargeting.} 
We adapt GMR~\cite{ze2025twist,joao2025gmr}, a real-time motion retargeting method, 
to the PICO human motion format (Figure~\ref{fig:VR human body}). 
The original GMR employs a two-stage optimization: (1) solving for link rotation consistency, and (2) refining global pose alignment. Since PICO motion capture often yields inaccurate global pose estimation, 
we modify the second optimization stage as follows: 1) for the lower body, optimize for position and rotation constraints; 2) for the upper body, only optimize for rotation constraints. This ensures 1) less feet sliding and 2) better upper-body teleportation experience.


We partition the retargeted links into lower-body $\mathcal{L}_{\mathrm{low}}$ (e.g., pelvis, hips, knees, ankles, feet) 
and upper-body $\mathcal{L}_{\mathrm{up}}$ (e.g., spine, shoulders, elbows, wrists, head). 
Let $R_i^{\text{human}}$ and $R_i^{\text{robot}}(\mathbf{q})$ be the link orientations, 
and $p_k^{\text{human}}$ and $p_k^{\text{robot}}(\mathbf{q})$ the link positions 
for a selected set of lower-body points $\mathcal{P}_{\mathrm{low}}$ (typically feet/ankles and optionally pelvis). 
To reduce sensitivity to noisy global pose estimation (and to support user teleportation), 
we measure all human positions in a pelvis-centric frame. 
The stage-2 optimization is then formulated as:
\begin{align}
\mathbf{q}^{*} = 
\arg\min_{\mathbf{q}} \; &
\sum_{i\in \mathcal{L}_{\mathrm{up}}\cup \mathcal{L}_{\mathrm{low}}}
w_i^{R}\,\big\|R_i^{\text{human}}-R_i^{\text{robot}}(\mathbf{q})\big\|_{F}^{2} 
\notag\\
&+\, \lambda_{\mathrm{pos}}
\sum_{k\in \mathcal{P}_{\mathrm{low}}}
w_k^{p}\,\big\|\,p_k^{\text{human,\,pelvis}}-p_k^{\text{robot}}(\mathbf{q})\,\big\|_{2}^{2}.
\label{eq:stage2}
\end{align}
Here $w_i^{R}$ and $w_k^{p}$ are per-link weights, 
$\lambda_{\mathrm{pos}}$ balances the rotation and position terms, 
and $p_k^{\text{human,\,pelvis}}$ denotes human keypoints expressed in the human pelvis frame. 
This formulation enforces accurate foot and ankle placement to mitigate foot sliding, 
while keeping the upper body free of positional terms so that 
global-pose jumps (e.g., teleportation) do not introduce artifacts—upper-body retargeting depends only on local rotations.

\noindent\textbf{Hand retargeting.} 
Directly mapping a human five-finger hand to the Unitree Dex31 hand is not intuitive for teleoperation, 
since the Dex31 only provides three fingers with limited degrees of freedom.
In practice, the functionality of the Dex31 hand is much closer to a parallel-jaw gripper than to a dexterous multi-fingered hand. 
Therefore, we simplify hand retargeting by treating the Dex31 as a gripper and not using hand pose estimation but controlling it by pressing buttons with PICO handheld controllers. 
We define two canonical configurations: an \textit{open pose} $\mathbf{q}_{\text{open}}$ and a \textit{close pose} $\mathbf{q}_{\text{close}}$. 
A scalar grasp command $\alpha \in [0,1]$ is computed from the human hand signals, where $\alpha=0$ denotes fully open and $\alpha=1$ denotes fully closed. 
The commanded Dex31 hand joint configuration is then interpolated as
\begin{equation}
\mathbf{q}_{\text{hand}} = (1 - \alpha)\,\mathbf{q}_{\text{open}} + \alpha \,\mathbf{q}_{\text{close}}.
\end{equation}
For tasks that require power grasp (\textit{e.g.}, grasp a cup) and tasks that require fine-grained pinching  (\textit{e.g.}, folding cloths), we define two sets of $\mathbf{q}_{\text{open}}$ and $\mathbf{q}_{\text{close}}$.

\begin{figure*}[t]
    \centering
    \includegraphics[width=1.0\linewidth]
    {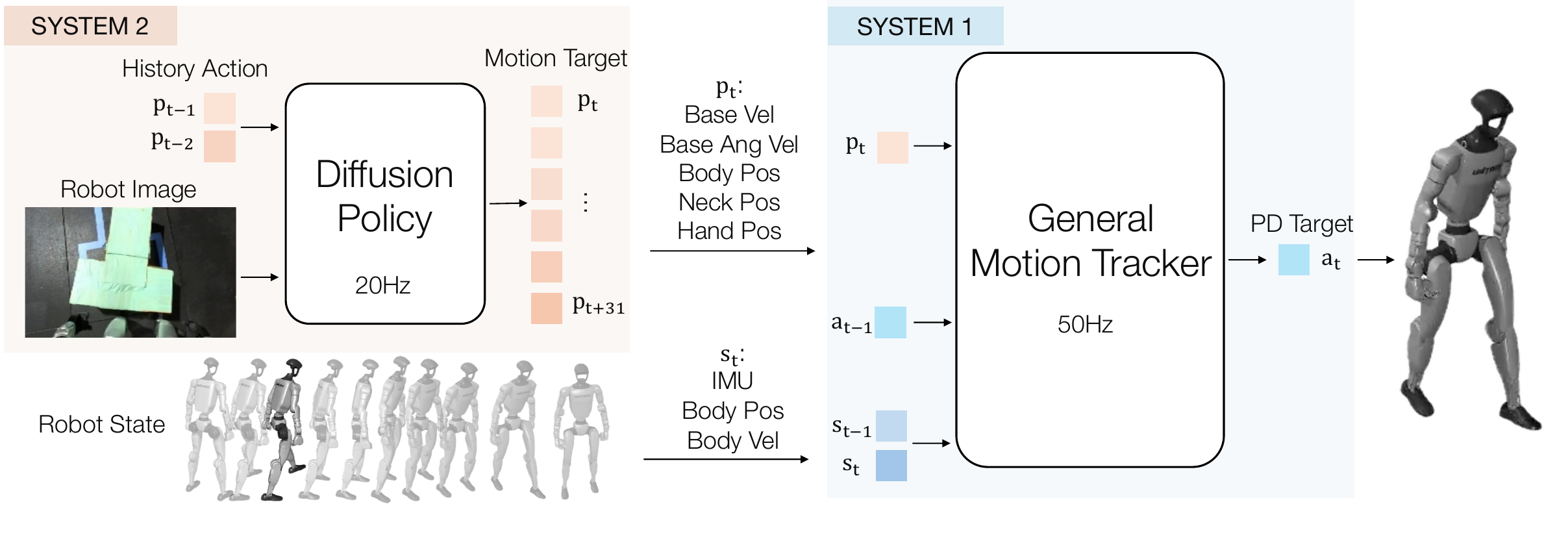}
    \caption{Hierarchical whole-body visuomotor policy learning framework built upon data collected via \ours. Unlike previous works that focus on upper-body manipulation or lower-body locomotion separately, our visuomotor policy controls the entire body, enabling complex tasks such as Kick-T that require coordinated whole-body movements.}
    \label{fig:policy}
    \vspace{-0.2in}
\end{figure*}
\noindent\textbf{Neck retargeting.} 
Let $R_{\text{head}}, R_{\text{spine}} \in SO(3)$ be the global rotations of the human head and spine in the world frame, respectively. The relative rotation is
\begin{equation}
R_{\text{rel}} = R_{\text{spine}}^{\top} R_{\text{head}}.
\end{equation}
From $R_{\text{rel}} = [r_{ij}]$, the robot neck joint targets are defined as
\begin{equation}
q_{\text{neck}}^{\text{yaw}} = \psi = \arctan2(r_{21},\, r_{11}),\,
q_{\text{neck}}^{\text{pitch}} = \theta = \arcsin(-r_{31}).
\end{equation}

\subsection{Training General Motion Trackers for Low-Level Control}
\label{sec:training controller}
To bring the retargeted kinematics motions onto a physical robot, we need a whole-body controller $\pi_\text{low}$ that takes into reference motions and outputs the desired PD target. Different from previous works that adopt a complex teacher-student pipeline to train a reasonable whole-body controller~\cite{ze2025twist,chen2025gmt,he2024omnih2o}, we design a simple one-stage training framework for general motion tracking. 

More specifically, we first curate a humanoid motion dataset consisting of around 20k motion clips. The motion dataset includes data retargeted via GMR~\cite{ze2025gmr,ze2025twist} (7k clips) and the original motion dataset from TWIST~\cite{ze2025twist} (13k clips). The motion data source includes AMASS~\cite{mahmood2019amass}, OMOMO~\cite{li2023omomo}, and our in-house MoCap data.  This mixture of the dataset ensures our policy learns omnidirectional walking. Similarly as found in TWIST~\cite{ze2025twist}, we find that curating a small set of motions from the teleoperation device is essential to bridge the domain gap. We only collect 73 motions via PICO, as these motions already cover most daily movements like walking, crouching, and manipulation. We then generate reward supervision from the motion datasets. The rewards are defined as $r=r_{\text{track}}+r_{\text{reg}}$, where $r_{\text{track}}$ is defined as:
\begin{equation}
    r_{\text{track}}=e^{-\alpha\|\mathbf{p}_{\text{cmd}}-\mathbf{p}_{\text{cur}}\|}
\end{equation}
where $\mathbf{p}_{\text{cur}}$ denotes the actual state the robot achieved. $r_{\text{reg}}$ consists of the regularization terms, such as the penalty on the action change.

The actor $\pi_{\text{low}}$ is trained via PPO and mainly consists of two parts: the convolutional history encoder and the MLP backbone. We find that compressing history robot proprioceptions and history reference motions into a compact latent vector boosts learning efficiency.

\begin{figure}[htbp]
    \centering
\includegraphics[width=1.0\linewidth]{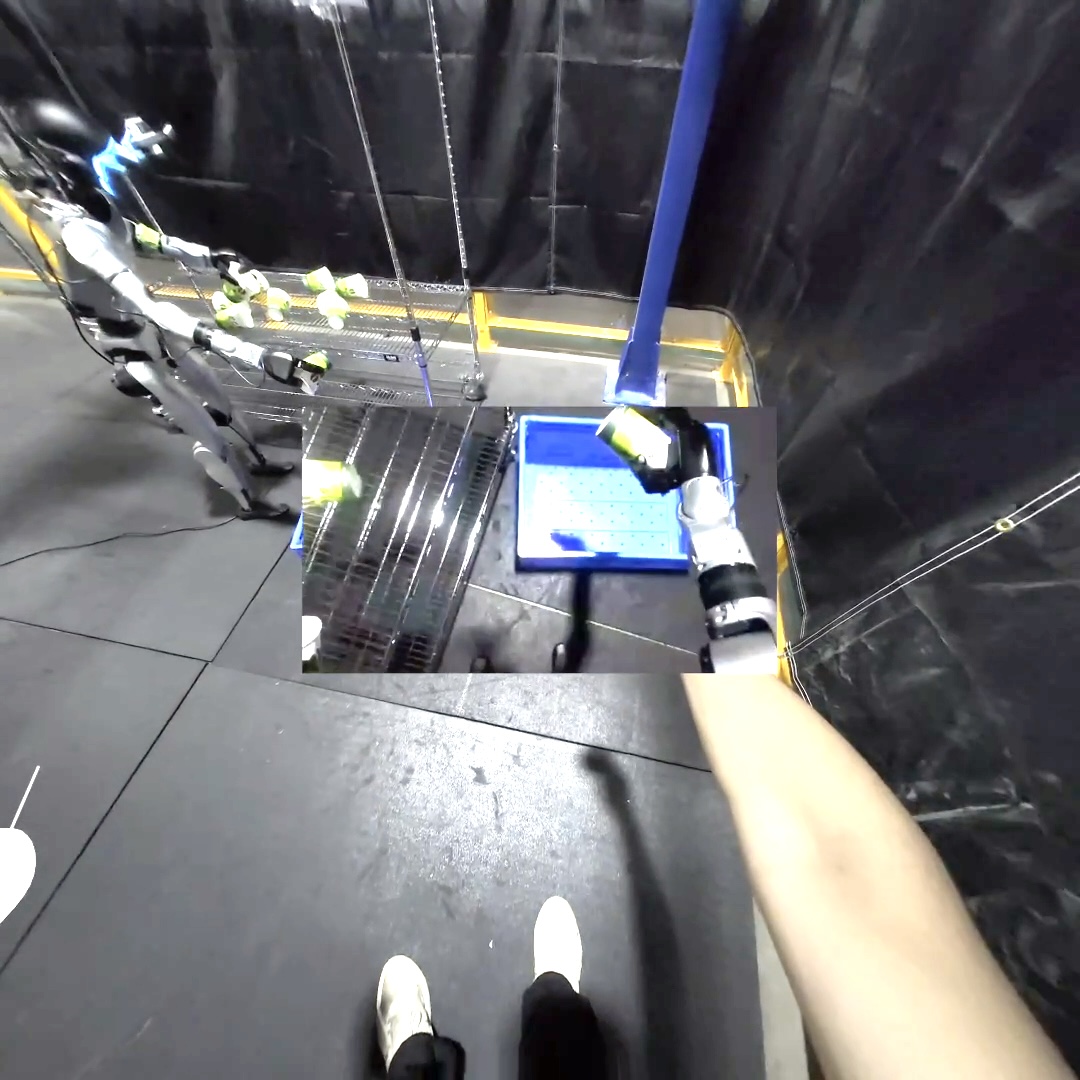}
    \caption{The view of the teleoperator in PICO. The robot vision is floating in the center.}
    \label{fig:egocentric teleop}
    \vspace{-0.2in}
\end{figure}

\subsection{Scalable Humanoid Data Collection}
\label{sec:scalable humanoid data collection}
We now describe our humanoid teleoperation and data collection system built with the aforementioned modules.

\noindent\textbf{Egocentric whole-body teleoperation.} During teleoperation, we obtain real-time streamed human motions from PICO (Section~\ref{sec:portable human data source}) and map human motions into robot motion commands $\mathbf{p}_{\text{cmd}}$,
and then send $\mathbf{p}_{\text{cmd}}$ to $\pi_{\text{low}}$ (Section~\ref{sec:training controller}) through Redis~\cite{redis_redis_2025}. Additionally, our teleoperation system is equipped with stereoscopic vision via the custom
shader implemented in ~\cite{zhao2025xrobotoolkit} that adjusts the interpupillary distance and sets the
focal point at approximately 3.3 feet, providing teleoperators with depth perception (see Figure~\ref{fig:egocentric teleop})
The stereo images are streamed from ZED Mini to PICO via GStreamer in the h265 format and to the data collection process via ZMQ in the JPEG format.

\begin{figure}[htbp]
    \centering
    \includegraphics[width=0.8\linewidth]{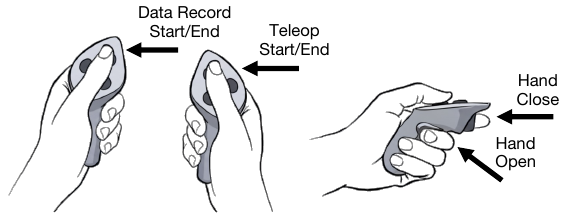}
    \caption{Illustrations on using the PICO joystick controller as the control center to make \ours a single-operator system.}
    \label{fig:pico controller}
    \vspace{-0.16in}
\end{figure}

\noindent\textbf{Single operator.} A practical teleoperation/data collection system should only require a single operator. Recent whole-body humanoid teleoperation systems focus on showing their capabilities~\cite{ze2025twist,li2025clone,li2025amo,lu2024mobiletv}, but most of them do not explicitly show how the teleoperation sessions start, pause, and terminate. AMO~\cite{li2025amo} and MobileTV~\cite{lu2024mobiletv} require two operators: one for the upper body and one for the lower body. TWIST~\cite{ze2025twist} and CLONE~\cite{li2025clone} require only one operator for teleoperating the robot, but need another one to control the start/end of the entire process. We program the PICO's handheld controllers to allow the demonstrator to safely and smoothly operate the entire system without the need for any assistance.
The handheld controllers play the role of the control center, as shown in Figure~\ref{fig:pico controller}.

\noindent\textbf{Safe control.} Humanoid robots are brittle; and this problem becomes more critical when designing a system that can fully control the robot. In \ours, we use motion interpolation for smooth state transition. For example, our system supports \textit{pause} via the origin joystick from PICO; and when \textit{pause} mode ends, we interpolate from the last robot pose to current target pose, to avoid sudden jump. This guarantees our system can operate in a quite long time safely and stop anytime when human operators are tired.

\noindent\textbf{System delay.} All modules in our system stream at a speed above $50$Hz, ensuring the overall delay to be lower than $0.1$s, significantly improved upon prior work \cite{ze2025twist} ($0.5$s delay).


\noindent\textbf{Data filtering.} During data collection, we consecutively record episodes. To process these trajectories, we developed a demonstration post-processing GUI that segments long sequences into multiple episodes, each corresponding to a completed task. We also reduce idle actions and remove failure episodes through filtering. 

\begin{figure*}[t]
    \centering
\includegraphics[width=1.0\textwidth]{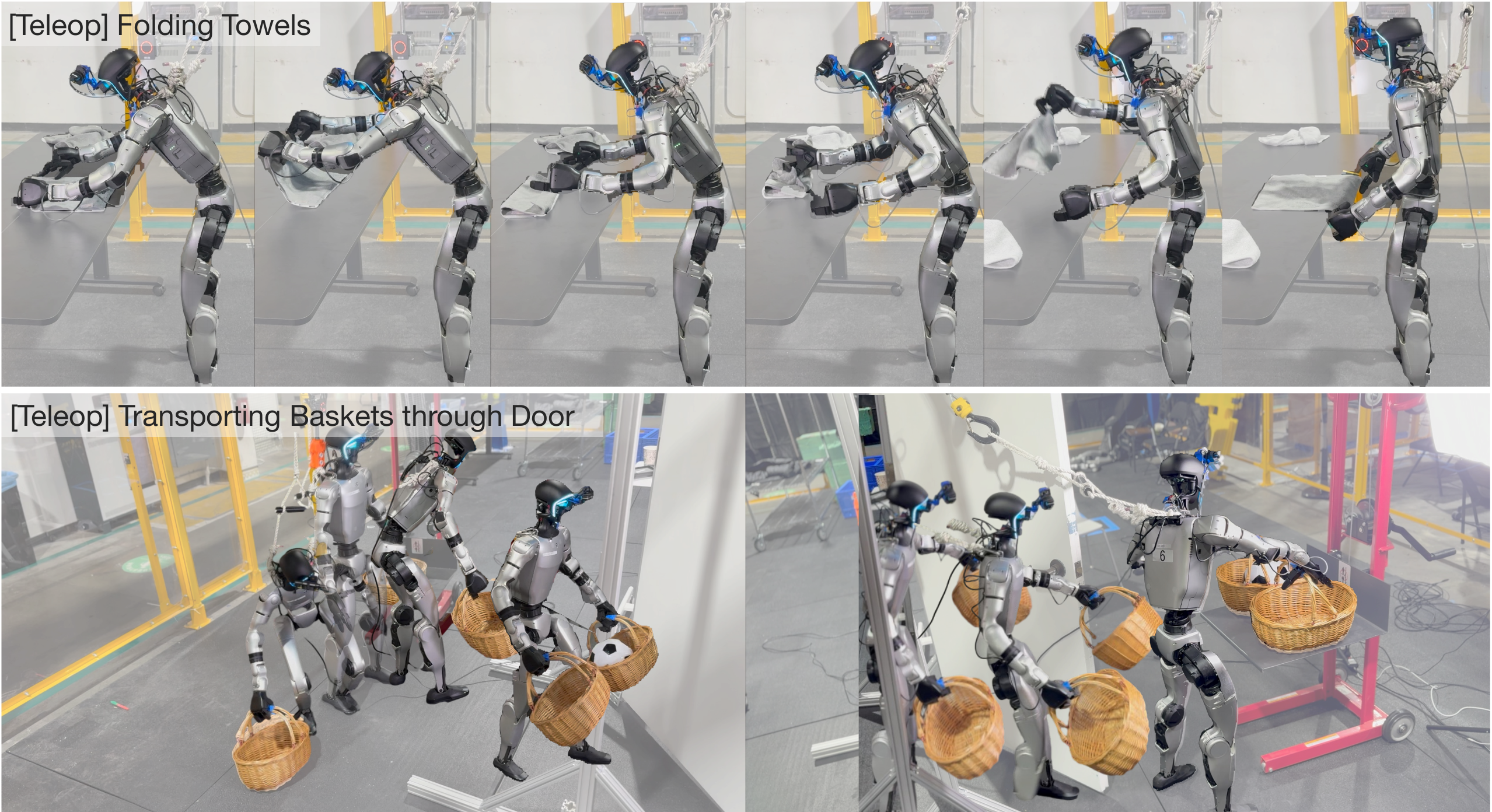}
    \caption{Long-horizon humanoid teleoperation powered by \ours. All tasks are achieved with streamed robot egocentric vision, full whole-body control, and a single operator.
    }
    \label{fig:long horizon teleop}
\vspace{-0.2in}
\end{figure*}

\subsection{Whole-Body Visuomotor Policy Learning}
\label{sec: whole-body policy learning}

Using the high-quality demonstration data collected through our teleoperation system, we develop a hierarchical visuomotor policy framework, as illustrated in Figure~\ref{fig:policy}. This section details the design and training of the high-level visuomotor policy $\pi_{\text{high}}^{\text{auto}}$.

\noindent\textbf{Observation and action space.} 
The visuomotor policy operates on visual observations and proprioceptive information to generate motion commands. Visual input consists of $360 \times 640$ RGB images captured by the ZED Mini camera, which are downsampled to $224 \times 224$ for computational efficiency. For robot proprioception, we use the historical command sequence $\mathbf{p}_{\text{cmd}}$ rather than raw robot states $\mathbf{s}$. This choice  of proprioception
serves two purposes: 1) it decouples the high-level policy from the low-level controller, enabling modular training and deployment, and 2) it mitigates error accumulation in this high-dimensional system by avoiding direct dependence on noisy raw robot states $s$.
The action space consists of the same command vector $\mathbf{p}_{\text{cmd}}$ used during teleoperation, ensuring consistency between data collection and policy execution. All proprioceptive inputs are normalized to improve training stability. 

\noindent\textbf{Network architecture.} 
We employ Diffusion Policy~\cite{chi2023diffusionpolicy} as our policy learning framework, utilizing 1D convolutional blocks for temporal modeling of action sequences. The policy predicts 64 action chunks using sample-based prediction~\cite{ze2024idp3,ze2024dp3}, corresponding to 2 seconds of future motion commands at the policy execution frequency. For visual encoding, we use a ResNet-18 backbone pre-trained with R3M~\cite{nair2022r3m}, which provides robust visual representations learned from diverse robotic datasets. 

\noindent\textbf{Data augmentation and regularization.} 
To enhance the robustness and generalization of the learned policy, we apply both state-space and visual augmentations. We inject 10\% Gaussian noise into the proprioceptive inputs, encouraging the policy to rely more heavily on visual observations rather than overfitting to precise state information. For visual augmentation, we employ a comprehensive set of techniques including random cropping, random rotation, and color jittering. These augmentations improve the policy's ability to generalize across different lighting conditions, camera viewpoints, and visual variations that may occur during deployment.

\noindent\textbf{Deployment and inference.} 
For efficient real-time execution, the trained Diffusion Policy is converted to ONNX format, achieving a 20Hz inference rate on a single NVIDIA RTX 4090. We execute 48 out of the predicted 64-step action chunks at 30Hz, maintaining consistency with the data collection frequency.

%% file: sec/4_exp.tex
\begin{figure*}[t]
    \centering
\includegraphics[width=1.0\textwidth]{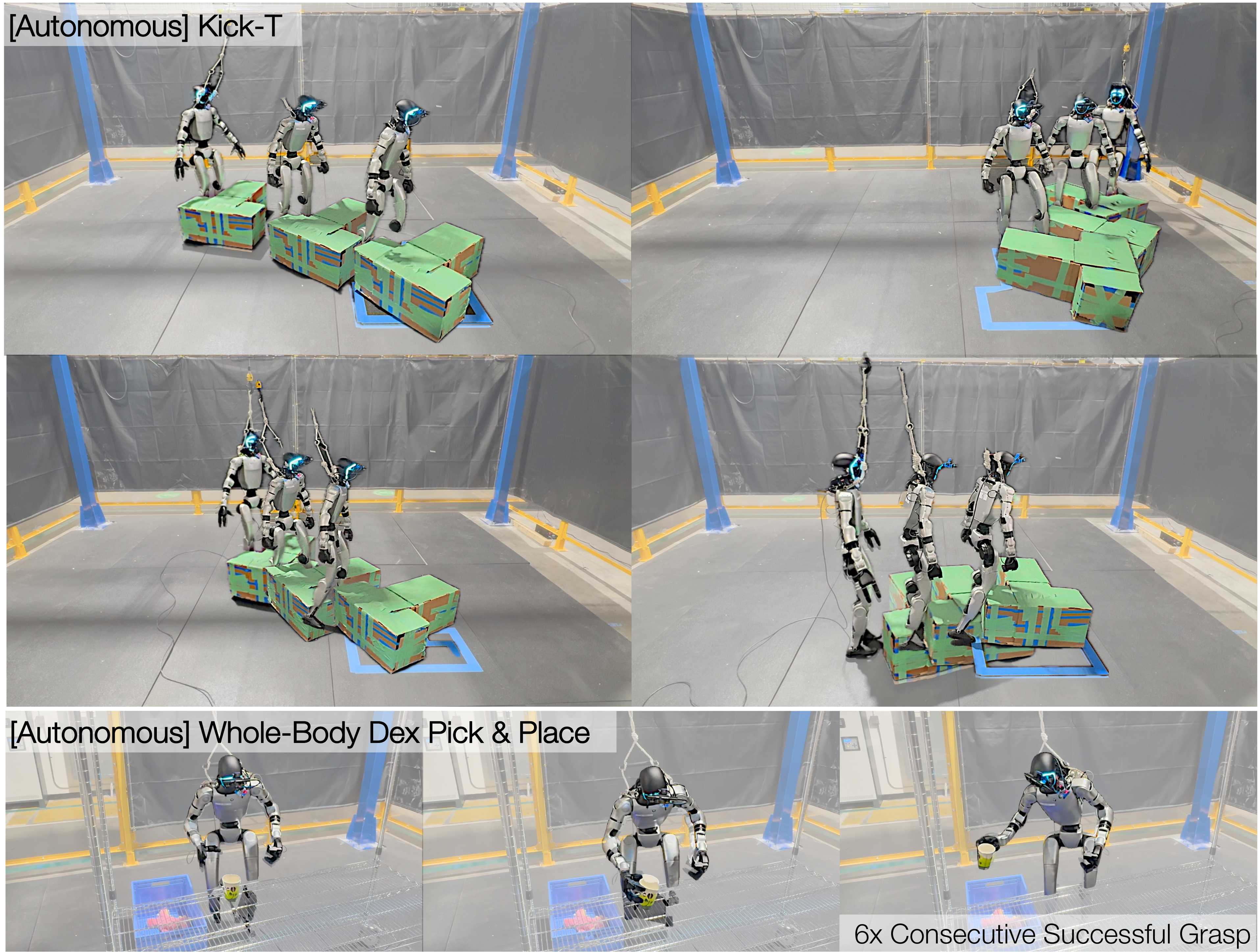}
    \caption{Closed-loop whole-body visuomotor policy execution in the real world. \ours enables effective and holistic whole-body humanoid data collection, which further enables versatile autonomous whole-body humanoid loco-manipulation \& legged manipulation skills.
    }
    \label{fig:policy roll out}
\vspace{-0.2in}
\end{figure*}
\section{Experiment Results}


In this section, we show that powered by \ours, we can 1) teleoperate Unitree G1 to perform long-horizon challenging whole-body dexterous tasks, 2) collect imitation learning data effectively, and 3) make Unitree G1 autonomously perform whole-body tasks via its egocentric vision.

\subsection{Long-Horizon Teleoperation}
\ours enables very long-horizon teleoperation. We showcase two representative tasks that cannot be achieved by previous systems (see Figure~\ref{fig:long horizon teleop}). We observe that 1) egocentric active perception and 2) smooth whole-body tracking instead of decoupled control are keys that enable such natural \& smooth, long-horizon, whole-body, and mobile tasks.

\noindent\textbf{Folding towels.} The robot uses its egocentric vision to locate the towel, move the towel to its front, grasp it, and shakes it to spread. Then it will pinch the corner to fold the towel in half with two hands. It repeats the motion to fold into thirds (or quarters) to the target size, presses along the crease to set it, and neatly places the finished towel to its left-hand side. The entire process requires fine-grained control of the wrists and hands, active vision, and whole-body reaching. Our robot can continuously fold 3 towels that are randomly placed on the table for now; and this is only bottlenecked by the underlying motor robustness, such as motor overheating.

\noindent\textbf{Transporting baskets through the door.} The robot first adjusts its position via changing foot placements and bends down to pick up the baskets on its left side and on its right side, respectively. We casually put the basket so the teleoperator seeks the basket first via robot active perception. Then the robot moves close to the door, pushes the door open with the arm, walks across the door, and places the basket gently onto the shelf. Note that all the base movements of the robot are achieved via a single teleoperator by tracking the lower-body movements.
\begin{figure}[t]
    \centering
   \includegraphics[width=1.0\linewidth]{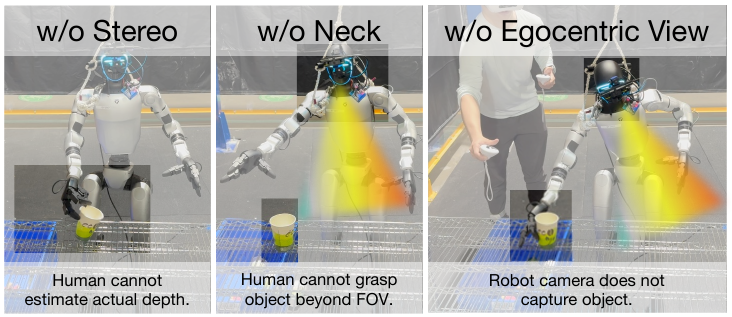}
    \caption{Comparison of different teleoperation settings. 
    }
    \label{fig:teleop comparison}
    \vspace{-0.2in}
\end{figure}

\subsection{Efficient Data Collection}
We show that 1) how effective \ours is in collecting imitation learning data and 2) how some key designs in our system improves data collection.

First, we show in Table~\ref{table:long data collection} that within 20 minutes, the expert teleoperator can consecutively collect 1) around 100 successful bimanual pick\&place or 2) around 50 successful mobile pick\&place.

\input{tables/long_data_collection}

Second, we conduct a user study to quantify the effectiveness of our data collection system. 
We evaluate two users: 1) an \textbf{expert} who has extensive experience using this system for data collection, and 2) a \textbf{novice} who is using the system for the first time during the test. 
Since the novice user gains proficiency through practice, we have them start with our complete system and then progressively remove features to isolate the impact of each component. 
As shown in Table~\ref{table:data_collection_efficiency}, \ours achieves the shortest completion times and highest success rates across all configurations.

As illustrated in Figure~\ref{fig:teleop comparison}, we observe several key findings: 1) without stereo vision for teleoperation, users tend to grasp higher than the actual object location, significantly increasing grasp failure rates; 2) without the neck module, users cannot perceive objects beyond the fixed field of view, making teleoperation extremely challenging; 3) when using third-person view with VR pass-through (\textit{i.e.}, \textit{w/o Egocentric View}), the expert can collect data remarkably fast (10 episodes in 43 seconds), but this is only possible because the expert stands directly beside the robot, which is infeasible for long-horizon mobile manipulation tasks which require remote control via egocentric vision.
\begin{figure}[htbp]
    \centering   \includegraphics[width=1.0\linewidth]{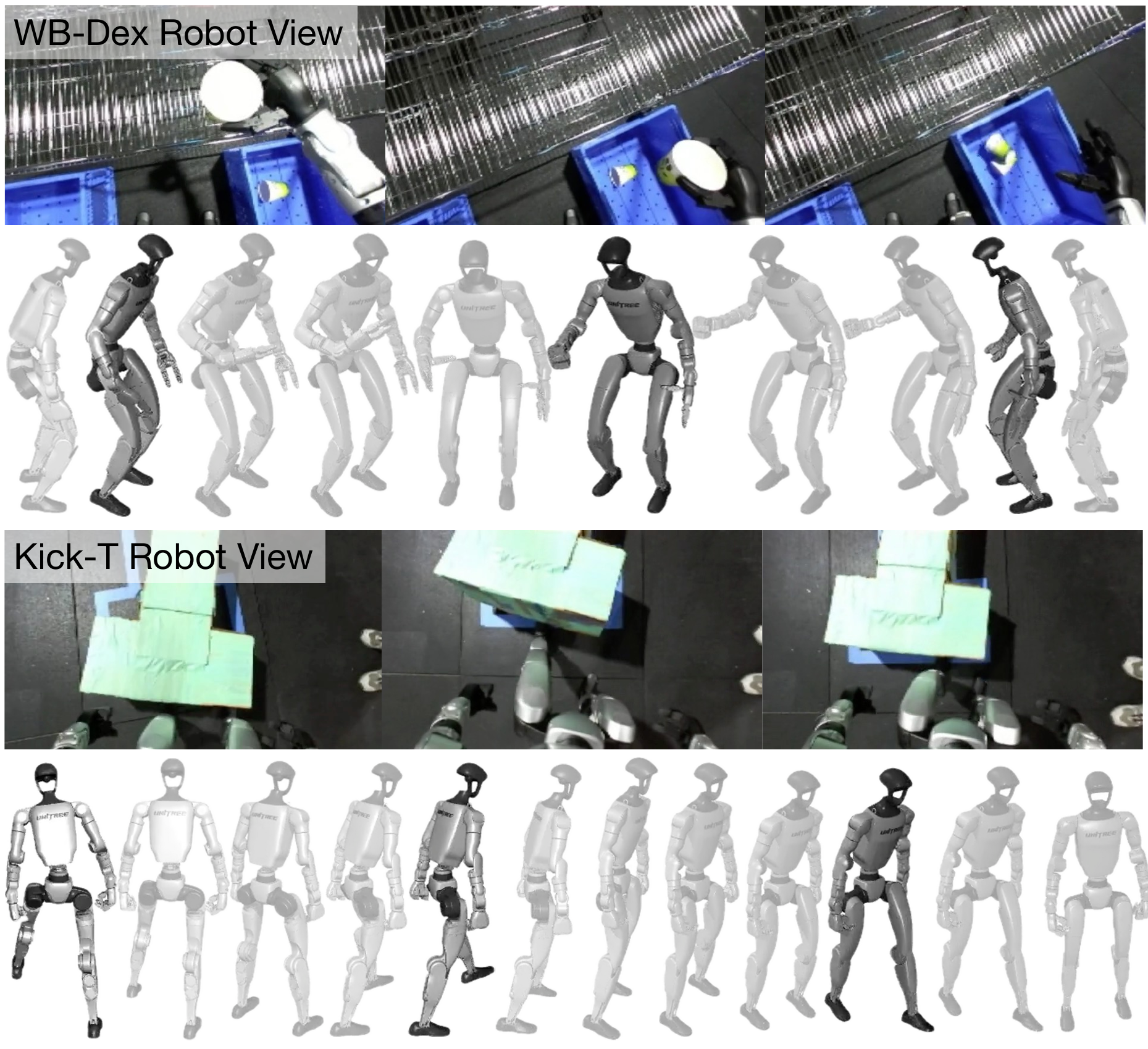}
    \caption{Visualization of training demonstrations (egocentric robot view and whole-body joint positions) for WB-Dex and Kick-T tasks. 
    }
    \label{fig:data vis}
    \vspace{-0.15in}
\end{figure}

\input{tables/data_collection_efficiency}

\subsection{Whole-Body Policy Learning Results}

We design two tasks to showcase autonomous results with our hierarchical visuomotor policy framework. We visualize the training data in Figure~\ref{fig:data vis}.

\begin{figure}[htbp]
    \centering
    \vspace{-0.2in}
    \includegraphics[width=1.0\linewidth]{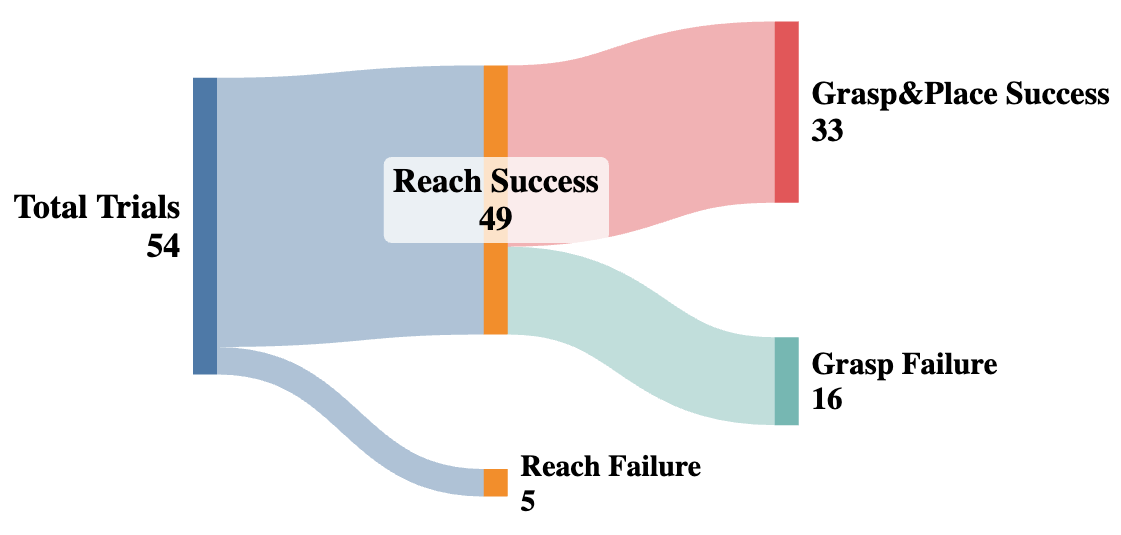}
    \vspace{-0.3in}
    \caption{All the success and failure cases in our WB-Dex task.}
    \label{fig:wbdex}
    \vspace{-0.1in}
\end{figure}

\noindent\textbf{Whole-body dexterous pick \& place (WB-Dex).} In this task, the robot bends down to pick up a cup from the shelf using its dexterous hand and places it into a box on the ground. We train the policy with 170 human demonstrations and report the success and failure rates in Figure~\ref{fig:wbdex}. We observe that the policy can reliably reach the cup in most cases. However, because the cup is very light, grasping it requires highly precise control; even a slight drift often results in grasp failure.

\noindent\textbf{Kick T-shaped box to target (Kick-T).} 
In this task, the robot uses its foot to kick a T-shaped green box toward a fixed T-shaped target position on the ground. The policy is trained with 50 demonstrations. In our data, the action pattern is consistent: the robot kicks with its left foot, and then takes a step forward with the right foot to maintain balance. This design ensures that the learned policy exhibits robust kicking behavior. We visualize policy rollouts in Figure~\ref{fig:policy roll out}. The policy successfully transports the T-shaped box to the target in 6 out of 7 trials. At present, the policy can only kick the box forward, without more flexible strategies such as walking around the box to adjust the kicking angle; we leave such capabilities to future work.

%% file: tables/long_data_collection.tex
\begin{table}[htbp]
\centering
\caption{Scalable data collection. We show that we can easily collect several demonstrations via our system.}
\label{table:long data collection}
\vspace{-0.05in}
\resizebox{0.48\textwidth}{!}{%
\begin{tabular}{l|cccc}
\toprule
Task & Time & \#Collected Episodes & Success Rate & Avg Time Per Episode\\
\midrule

 Bimanul Manip 1  & 18.5 min & 98 & 100\% & 11 s\\

Mobile Manip & 19.5 min & 46 & 100\%  & 25 s \\
\bottomrule
\end{tabular}}
\vspace{-0.1in}
\end{table}

%% file: tables/data_collection_efficiency.tex

\begin{table}[h]
\centering
\caption{Data collection efficiency of \ours on different users and setups. The results show the necessity of using active egocentric stereo vision.}
\label{table:data_collection_efficiency}
\vspace{-0.05in}
\resizebox{0.5\textwidth}{!}{%
\begin{tabular}{l|ccc|S[table-format=3.1]|S[table-format=3.1]|S[table-format=3.1]}
\toprule
\textbf{Collect 10 Demos} 
& \multicolumn{3}{c|}{\textbf{Succes/Total Trials}} 
& \multicolumn{3}{c}{\textbf{Time Cost (s)}} \\
\cmidrule(lr){2-4} \cmidrule(lr){5-7}
& Novice & Expert & Avg (Sum) & {Novice} & {Expert} & {Avg (Mean)} \\
\midrule
\ours            & 10/12 & 10/11 & \textbf{20/23} & 75.6 & 59.9 & 
\textbf{67.8}\\
w/o Stereo       & 10/12 & 10/15 & 20/27           & 90.6 & 105.9 & 98.3 \\
w/o Neck         & 7/17  & 9/12  & 16/29           & 144.0 & 80.5 & 112.3 \\
w/o Egocentric View & 10/13 & 10/10 & \textbf{20/23} & 94.3 & 43.0 & 68.7 \\
\bottomrule
\end{tabular}}
\vspace{-0.1in}
\end{table}





%% file: sec/5_conclu.tex
\section{Conclusions and Limitations}

We introduce \ours, a portable and holistic mocap-free data collection system for humanoid robots with full whole-body control. By combining lightweight VR devices with an attachable neck for egocentric vision, our framework enables scalable data collection. On top of this, we designed a hierarchical visuomotor policy that allows a real humanoid robot to autonomously perform versatile whole-body skills including whole-body dexterous manipulation and Kick-T.

\noindent\textbf{Limitations.} 1) The general motion tracker struggles with highly dynamic movements such as sprinting due to challenges in tracking fast, complex motions. 2) PICO's whole-body pose estimation is less accurate than high-cost motion capture systems, particularly for elbows and knees where no trackers are placed, resulting in reduced motion quality.

\section{Discussions on Scaling Up Humanoid Data}
There are several key challenges that must be addressed before we can realistically scale up high-quality humanoid robot data.




\noindent\textbf{Standardizing humanoid hardware.} There is currently a wide range of humanoid platforms available for research—such as the Unitree G1/R1 and Booster T1/K1—which makes collected data difficult to reuse due to cross-embodiment discrepancies. Recently, the Unitree G1 has emerged as a popular choice because it offers a strong balance between performance and cost. At this stage, we advocate for standardizing humanoid hardware usage before attempting to scale data collection. In our work, we adopt the Unitree G1 platform and highlight that egocentric vision is crucial for capturing human-level manipulation data. We therefore introduce a low-cost neck add-on for the G1 that enables an egocentric camera setup and delivers roughly 80\% of core human functionality.

\noindent\textbf{Democratizing humanoid data collection.}
High-quality humanoid datasets have traditionally relied on motion-capture systems, limiting data collection to MoCap studios and specialized equipment. We demonstrate a portable and cost-efficient alternative that preserves rich human manipulation capabilities rather than only enabling whole-body movement.

\noindent\textbf{Sharing humanoid data.}
We believe that open humanoid datasets should serve as a foundation for future research. To that end, we publicly release all collected humanoid data on HuggingFace, and provide visualizations at \href{https://twist-data.github.io}{https://twist-data.github.io}
, with the goal of making our dataset directly reusable and easily extendable by the community.

\begin{figure}[t]
    \centering
   \includegraphics[width=1.0\linewidth]{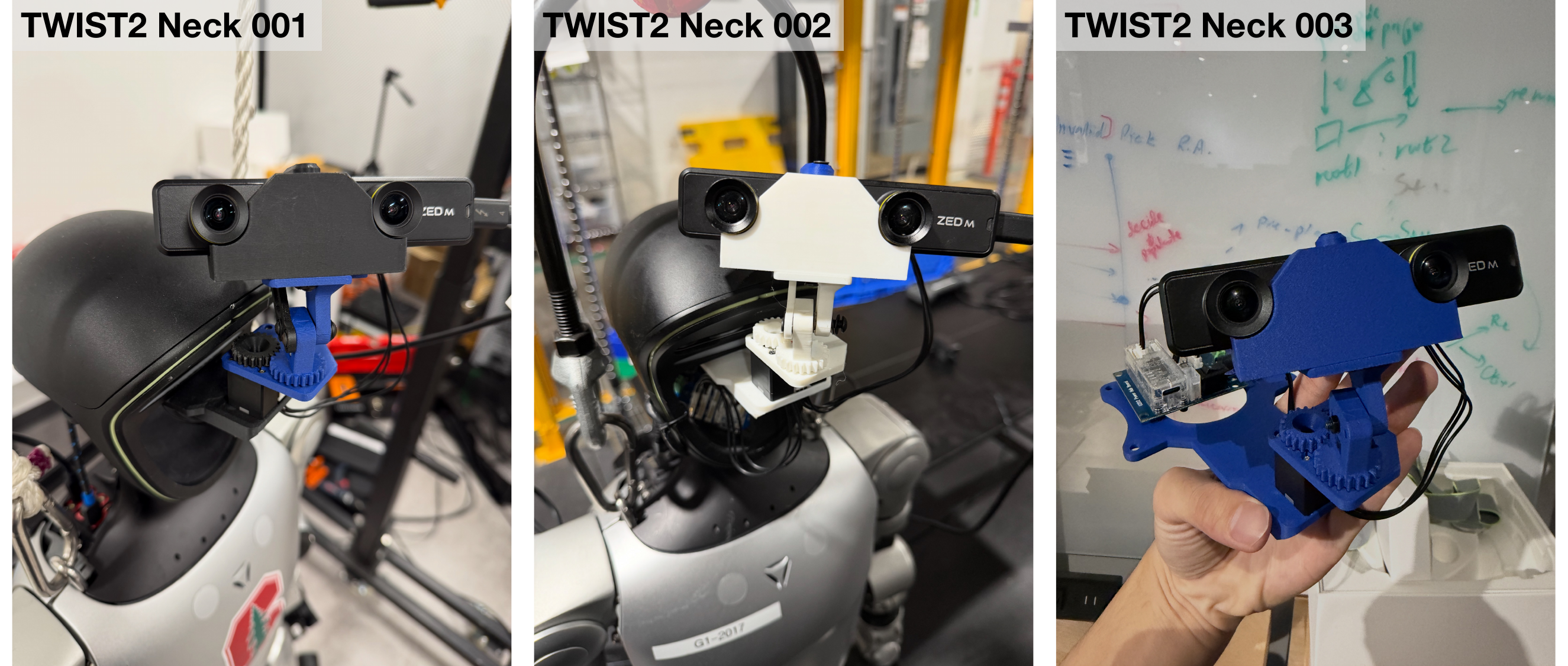}
    \caption{We have manufactured 3 TWIST2 Necks, indicating that TWIST2 Neck is easy to assemble and can be democratized for research purposes.}
    \label{fig:twist2 neck 123}
    \vspace{-0.1in}
\end{figure}

\section*{Acknowledgments}
We want to thank Charlie Cheng, Shaofeng Yin, Yuanhang Zhang, Yunchu Zhang, and Raven Huang for their help in real-world experiments. We also thank Wenhao Wang, Ke Jing, Ning Yang, Liuchuan Yu, and Zhigen Zhao
 for the helpful discussion in the PICO usage. The human motion datasets used in this work, including AMASS~\cite{mahmood2019amass} and OMOMO~\cite{li2023omomo}, are solely for research purposes.